\documentclass{article}
\usepackage{kr}

\usepackage{times}
\usepackage{soul}
\usepackage{url}
\usepackage[hidelinks]{hyperref}
\usepackage[utf8]{inputenc}
\usepackage[small]{caption}
\usepackage{graphicx}
\usepackage{amsmath}
\usepackage{amsthm}
\usepackage{booktabs}
\usepackage{algorithm}
\usepackage{algorithmic}
\usepackage{mathtools}

\usepackage{multirow}
\usepackage{amsmath}

\title{GPT-based Textile Pilling Classification Using 3D Point Cloud Data}

\author{%
Yu Lu$^{1,2,*}$\and
YuYu Chen$^{1,2}$\and
Gang Zhou$^1$\and
Zhenghua Lan$^{1}$\\
\affiliations
$^1$Hangzhou Innovation Institute, Beihang University\\
$^2$State Key Laboratory of Virtual Reality Technology and Systems, Beihang University, Beijing, China\\
\emails
\{yulucas, yuyu\_chen, zhenghua\}@buaa.edu.cn,
zhougang@bhhzcxyjy.wecom.work
}

\begin{document}

\maketitle

\begin{abstract}
  Textile pilling assessment is critical for textile quality control. We collect thousands of 3D point cloud images in the actual test environment of textiles and organize and label them as TextileNet8 dataset. To the best of our knowledge, it is the first publicly available eight-categories 3D point cloud dataset in the field of textile pilling assessment. Based on PointGPT, the GPT-like big model of point cloud analysis, we incorporate the global features of the input point cloud extracted from the non-parametric network into it, thus proposing the PointGPT+NN model. Using TextileNet8 as a benchmark, the experimental results show that the proposed PointGPT+NN model achieves an overall accuracy (OA) of 91.8\% and a mean per-class accuracy (mAcc) of 92.2\%. Test results on other publicly available datasets also validate the competitive performance of the proposed PointGPT+NN model. The proposed TextileNet8 dataset will be publicly available.
\end{abstract}

\section{Introduction}

Textiles play an important role in our lives. Pilling, a common issue associated with textiles, typically occurs when textiles are pulled out or twisted into ball shapes due to friction or movement \cite{a0liu,a1jian2022,a2zhang}, which not only damages the appearance of textiles but also significantly impacts their tactile sensation and mechanical properties. The assessment of textile pilling involves employing specialized equipment to polish textiles followed by visual inspection or automatic methodologies to categorize them based on pill quantity and characteristics \cite{a3palmer,a4chen}. The results of pilling assessment serve as critical indicators of textile quality and are of great concern to textile manufacturers. According to the industry consensus, textile pilling is typically categorized into multiple grades contingent upon the severity of the pilling phenomenon \cite{a5furferi2014towards}.

Traditional manual evaluation methods rely on human experience, which is subjective and time-consuming. Several studies \cite{a4chen,a6zhang,a7yap,a5furferi2014towards,a0liu,a1jian2022} have proposed algorithmic and automated equipment-based approaches for textile pilling assessment, most of which utilize 2D cameras to capture images of the textiles. These methods employ various image-processing algorithms to detect the quantity and characteristics of pilling and then grade the textiles. However, these methods often rely on proprietary and non-public datasets that are small in size, and analysis based on 2D images is susceptible to environmental factors, making it challenging to effectively validate the generality of the proposed methods.

\begin{figure}[t]
   \centering
   \includegraphics[width=0.93\linewidth]{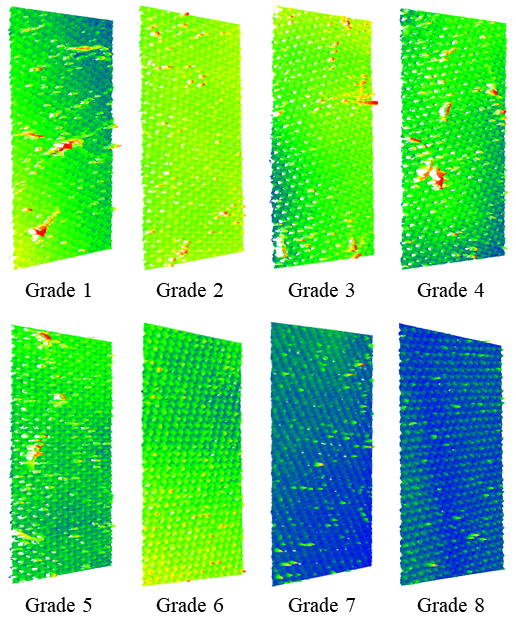}
   \caption{Sample images from the collected point cloud data of polyester. Best viewed in color.}
   \label{fig:fig1}
\end{figure}

3D point cloud data has demonstrated its extensive applications across various fields, including medical engineering \cite{b3pl,b1chrepeat,b2krawczyk2023segmentation}, architectural engineering \cite{b4bou,b5wang,b6xu2023approach}, and autonomous driving \cite{b8moon,b72020deep6,b9chen}. While point cloud data has been utilized in the textile industry, there is limited research on the evaluation of textile pilling. Compared to image data acquired with 2D cameras, point cloud data acquired with 3D cameras is less susceptible to environmental changes and directly reflects the object's shape and structure, offering detailed geometric information. These advantages contribute to the widespread usage of 3D point cloud data in object recognition and measurement.

In this study, we exploit 3D point cloud data for textile pilling classification. We collected thousands of point cloud images of textiles and organized them into the TextileNet8 dataset comprising eight categories after data cleaning and annotation. Figure~1 shows an example of the collected raw point cloud images. The PointGPT model \cite{b10chen2024pointgpt} utilizes a point cloud sequencer and a dual mask strategy to perform tasks such as semantic segmentation and classification on point cloud data, yielding promising results on publicly available datasets. We developed an improved version of the PointGPT model and applied it to classify textile pilling, obtaining satisfactory results on the TextileNet8 dataset.

In summary, our contributions are as follows. First, we construct an eight-categories dataset for textile pilling called TextileNet8, which, to the best of our knowledge, is the first publicly available eight-categories point cloud dataset in the field of textile pilling evaluation and can be used as a benchmark for textile pilling assessment. We evaluate the performance of various point cloud classification models on the proposed dataset and benchmark the textile piling classification task. Second, we introduce an improved model for fusing the global information of the point cloud, which is named PointGPT+NN. The proposed model achieves an overall accuracy (OA) of 91.8\% and a mean accuracy per class (mAcc) of 92.2\% on the proposed dataset.

\begin{figure*}[t]
   \centering
   \includegraphics[width=0.85\linewidth]{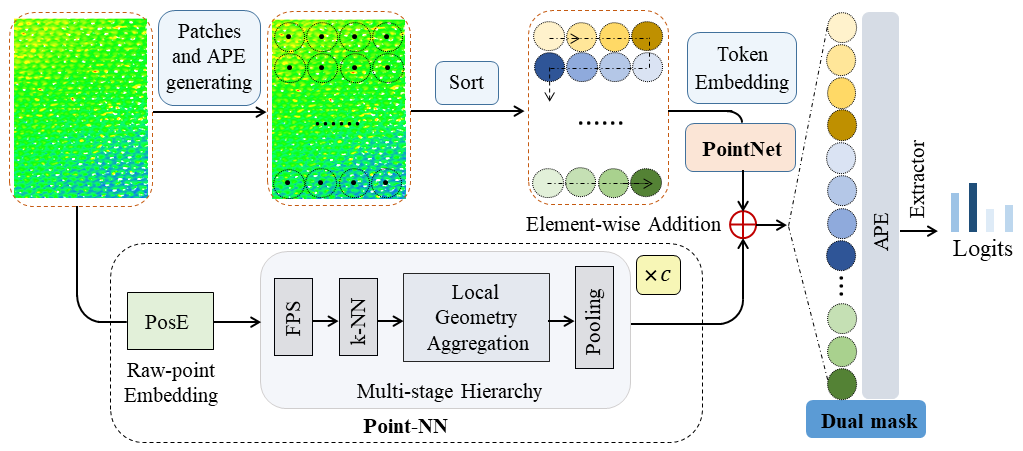}
   \caption{Overall architecture of the proposed PointGPT+NN. The input point cloud image is divided into multiple point patches, which are then sorted and arranged in an ordered sequence. The absolute position encoding (APE) is also generated in this stage. The arranged point patches are then put into a PointNet network for token embedding. Meanwhile, the input point cloud image is fed into a Point-NN network to do feature embedding on a complete point cloud image. $c$ in the Point-NN network denotes the stage number of the multi-stage hierarchy. The features generated by the PointNet and the Point-NN are then fused by element-wise addition. The fused features and the APE are then put into the extractor, while a dual mask strategy is applied to generate the logit classification results. Best viewed in color.}
   \label{fig:fig2}
\end{figure*}

\section{Related Work}

\subsection{Textile Pilling Evaluation}

The majority of automated textile pilling assessment methods utilize 2D images. Zhang et al. \shortcite{a2zhang} proposed a method combining 2DDTCWT image reconstruction with multi-layer perceptrons (MLP) classification and conducted experiments on 203 self-made textile pilling images to evaluate its effectiveness. Furferi et al. \shortcite{a5furferi2014towards} developed a machine vision-based program to extract fabric parameters, subsequently training an artificial neural network for automatic textile pilling classification using the extracted features. Xiao et al. \shortcite{b11xiao2021objective} utilized a series of image analysis techniques, including the Fourier transform, multidimensional discrete wavelet transform, and an iterative thresholding method, followed by deep learning classification for textile pilling objective evaluation. Yap et al. \shortcite{a7yap} employed a support vector machine with a radial basis function kernel for wool knitwear pilling evaluation, utilizing 17 collected textile pilling features.

Analysis based on 2D images is susceptible to environmental factors like illumination variation. Some researchers leverage 3D information in textile pilling assessment to enhance the reliability of objective evaluation methods. Jian et al. \shortcite{a1jian2022} proposed a semi-calibrated near-light photometric stereo (PS) method, employing the PS algorithm for 3D depth retrieval from 2D images, followed by segmentation and classification using global iterative thresholding and k-nearest neighbors (KNN) methods. Liu et al. \shortcite{a0liu} utilized structure from motion (SFM) and patch-based multi-view stereo (PMVS) algorithms for depth acquisition via 3D reconstruction, subsequently employing edge detection, adaptive threshold analysis, and morphological analysis for pilling segmentation and classification. Ouyang et al. \shortcite{b12ouyang2013fabric} obtained depth information via 3D reconstruction, utilizing depth and distance criteria for seed selection and region growth to characterize pilling appearance. Fan et al. \shortcite{b13fan} proposed a textile pilling rating assessment system integrating active contour and neural networks, which utilizes a multi-view stereo vision algorithm for fabric surface reconstruction. The above 3D information-based textile pilling assessment methods still use 2D images and rely on 3D reconstruction to obtain more information about textile pilling.

\subsection{3D Point Cloud Classification}

Graph-based methods for point cloud classification involve transforming point cloud data into a graph structure. The Dynamic Graph CNN (DGCNN) \cite{c1wang2019dynamic} model dynamically computes graphs at each network layer and introduces a novel neural network module called EdgeConv. This module captures local neighborhood information of the point cloud while learning global shape features. Li et al. \shortcite{c2li2019deepgcns} introduced a new method for training deep graph convolutional networks (GCNs), leveraging residuals, dense connectivity, and dilated convolutions from convolutional neural networks (CNNs) adapted to the GCN architecture. They constructed a 56-layer DeepGCN model using these techniques. Xu et al. \shortcite{c3xu2020grid} proposed a method for rapid and scalable point cloud learning named Grid-GCN. Grid-GCN employs a novel data structuring strategy known as coverage-aware grid query (CAGQ), which enhances spatial coverage and reduces time complexity by leveraging the efficiency of the grid space. Hu et al. \shortcite{c4hu2023m} introduced a multi-scale graph convolutional network (M-GCN) that extracts local geometric features based on multi-scale feature fusion. This method effectively enriches the representation capability of point clouds by extracting local topological information across scales.

Point cloud classification methods based on MLP generally independently process features for each point. Qi et al. \shortcite{c5qi2017pointnet} introduced PointNet, a novel neural network that can directly process point cloud data. The model preserves the alignment invariance of the point cloud and provides a unified architecture for point cloud classification, segmentation, and semantic parsing of scenes for the first time. Based on PointNet, Qi et al. \shortcite{c6qi2017et++} proposed PointNet++, which can learn local features at an incremental contextual scale by recursively applying PointNet to the nested segmentation of point clouds. PointNet++ improves the ability to recognize fine-grained patterns and generalize to complex scenes. Qian et al. \shortcite{c7qian} reviewed the previous works, optimized the training and model expansion strategies, introduced an inverted residual bottleneck design and separable MLPs for efficient and effective model scaling, and introduced PointNeXt. Zhang et al. \shortcite{c8zhang2023} constructed a non-parametric network (Point-NN) based on furthest point sampling (FPS), KNN, and pooling operations with trigonometric functions. They also introduced a parametric network (Point-PN) by inserting linear layers on the Point-NN.

Following the success of Transformer models in NLP and image processing, researchers have begun exploring Transformer-based approaches for point cloud processing. Among these approaches, PointGPT stands out due to its innovative dual mask strategy and the incorporation of distinctive extractor-generator transformer architectures. PointGPT has demonstrated promising results on publicly available 3D point cloud analysis datasets.

\section{Methodology}

\subsection{Generative Pre-training Transformer(GPT) for Point Cloud}

The generative pre-training transformer (GPT) \cite{c9radford2018improving} excels at learning representative features, primarily through auto-regressive prediction. PointGPT utilizes the GPT framework for analyzing point clouds, which involves two main stages: sorting and embedding, and auto-regression pre-training. In the sorting and embedding stage, $n$ center points are selected by FPS, followed by KNN to construct $n$ point patches around each center point. These center points are sorted using Morton code \cite{d1wei2020weighted}, and the corresponding point patches are arranged accordingly. The sorted point patches ($P^s$) are put into a PointNet network \cite{c5qi2017pointnet} to extract geometric information for feature embedding. The embedding of sorted point patches is as follows \cite{b10chen2024pointgpt}: 
\begin{equation}
T=\mathrm{PointNet}(P^s).
\tag{1}
\label{eq:eq1}
\end{equation}%
where $T$ represents the embedding tokens with D-dimensional.

In the auto-regression pre-training stage, PointGPT employs a dual mask strategy to enhance the overall understanding of the point cloud compared to other mask methods. Specifically, the other mask methods mask half of the input token, while the dual mask strategy randomly masks an additional portion of the input token on top of the vanilla mask. The self-attention process with the dual mask strategy is as follows \cite{b10chen2024pointgpt}:
\begin{equation}
    \mathrm{Selfattention}(T)=\mathrm{softmax}(\frac{Q K^{T}}{\sqrt{D}}-(1-M^d)\cdot\infty)V.
\tag{2}
\label{eq:eq2}
\end{equation}%
where $Q$, $K$, $V$ are $T$ encoded with different weights for the $D$ channels, and $M^d$ is the dual mask which is 0 or 1 on the location.

The extractor in PointGPT consists only of transformer decoder blocks that extract potential representations using a dual mask strategy. In addition, sinusoidal position encoding (PE) is applied to map the center points to absolute position encoding (APE). The generator in PointGPT also consists of transformer blocks responsible for generating point tokens for the subsequent prediction head. The directions relative to the next point patches are provided in the generator to maintain consistency in the order of the point patches.

\begin{table*}[t]
\begin{center}
\begin{tabular}{c|cc|ccccccccc|c}
\toprule
                  & \multicolumn{2}{c|}{Attributes}                      & \multicolumn{9}{c|}{Grades}                                                             & \multirow{2}{*}{\begin{tabular}[c]{@{}c@{}}Total \\ Num.\end{tabular}} \\ \cmidrule{2-12}
                  & \multicolumn{1}{c|}{Public Access} & Image Type      & 1 & 2 & 3 & 4 & 5 & 6 & 7 & 8 & 9 &                                                                        \\ \midrule
Luo et al. \shortcite{a1jian2022}        & \multicolumn{1}{c|}{No}            & 2D images       & 11      & 18      & 20      & 30      & 19      & /       & /       & /       & /       & 98                                                                     \\
Liu et al. \shortcite{a0liu}        & \multicolumn{1}{c|}{No}            & 2D images       & 16      & 27      & 20      & 22      & 20      & /       & /       & /       & /       & 105                                                                    \\
Zhang et al. \shortcite{a6zhang}      & \multicolumn{1}{c|}{No}            & 2D images       & 32      & 63      & 46      & 49      & 13      & /       & /       & /       & /       & 203                                                                    \\
Furferi et al. \shortcite{a5furferi2014towards}    & \multicolumn{1}{c|}{No}            & 2D images       & 15      & 15      & 15      & 15      & 15      & 15      & 15      & 15      & 15      & 135                                                                    \\ \midrule
TextileNet8 (Ours) & \multicolumn{1}{c|}{Yes}           & 3D point clouds & 150     & 165     & 180     & 165     & 165     & 165     & 180     & 165     & /       & 1335                                                                   \\ \bottomrule
\end{tabular}
\caption{Statistics of TextileNet8 and comparison to other datasets.} \label{tab:tab1}
\end{center}
\end{table*}

The final component of the auto-regressive pre-training stage is a prediction head, which is responsible for predicting subsequent point patches. The prediction head comprises a two-layer MLP with fully connected (FC) layers and rectified linear unit (ReLU) activation. The prediction head maps tokens generated by the generator to vectors, subsequently restructured to construct the predicted point patches.

As for the loss function in PointGPT, the generation loss $L^g$ uses the $l_{1}$-form and $l_{2}$-form of the Chamfer distance (CD) \cite{d2fan2017point}, denoted as $L^g_{1}$ and $L^g_{2}$, and is computed as $L^g$ = $L^g_{1}$+$L^g_{2}$. Specifically, the following objective: $L^{f}$ = $L^{d}$+$\gamma$$L^{g}$ is optimized during the fine-tuning stage, where $L^{d}$ represents the loss for the downstream task, which is CrossEntropyLoss \cite{d3zhang2018generalized} for the classification task. $\gamma$ balances the contribution of the loss of the downstream task and the loss of the generation task.

\subsection{Construction of PointGPT+NN}

Based on a comprehensive understanding of PointGPT, we identify a potential limitation in its feature embedding process for textile pilling grading purposes. Specifically, while PointGPT utilizes the PointNet network for feature embedding of each point patch, it fails to conduct feature embedding on a complete point cloud image. To address this, we propose a method to fuse features of the entire input point cloud during feature embedding.

One promising solution is the recently developed Point-NN \cite{c8zhang2023}, a non-parametric network tailored for 3D point cloud analysis. We combined the global features of the input point cloud extracted by the Point-NN model with PointGPT to find a way to enhance the performance of PointGPT on the textile pilling assessment task. Figure~2 shows the overall architecture of the constructed PointGPT+NN model.

Point-NN comprises a non-parametric encoder (NPE) for 3D feature extraction and a point memory bank (PMB) for task-specific recognition. For this study, we only use NPE for 3D feature extraction purposes. As shown in Figure 2, we only utilize the NPE for feature extraction purposes. The NPE first uses trigonometric functions for the positional encoding (PosE) and extends it for non-parametric 3D embedding. Sine and cosine trigonometric functions are used to embed a raw point into a vector \cite{c8zhang2023}: 
\begin{equation}
\begin{split}
    & f_i^x[2m]=\mathrm{sine}\left(\alpha x_i/\beta^{\frac{6m}{C_{I}}}\right), \\
    & f_i^x[2m+1]=\mathrm{cosine}\left(\alpha x_i/\beta^{\frac{6m}{C_{I}}}\right),
\end{split}
\tag{3}
\label{eq:eq3}
\end{equation}%
where $\alpha$ and $\beta$ control the magnitude and wavelengths, respectively. $C_{I}$ denotes the initial feature dimension, and $m\in\left[0,{\frac{C_{I}}{6}}\right]$ is the channel index. The NPE then adopts a multi-stage hierarchy structure, leveraging farthest point sampling (FPS), KNN, trigonometric function-based local geometry extraction, and pooling for incremental aggregation of local geometry.

This multi-stage hierarchy structure design of NPE aims to generate high-dimensional global features for the point cloud. It is worth noting that Point-NN employs simple trigonometric functions to capture local spatial geometric information and fine-grained semantics of various 3D structures without any learnable operators. When extracting global features of the input point cloud images using NPE, the number of stages in the multi-stage hierarchy structure can be varied to control the complexity of the features. In addition, adjusting the values of $\alpha$ and $\beta$ can better extract the positional relationships of the points.

As shown in Figure 2, the extractor in PointGPT is utilized to enhance the semantic level of the latent representations learned in the pre-training stage. The feature extracted by NPE of Point-NN is fused with PointGPT by element-wise addition as follows:

%
\begin{equation} 
    T=\text {Point-NN}\left(P^{r}\right)+\lambda \times \operatorname{PointNet}\left(P^{s}\right)
\tag{4}
\label{eq:eq4}
\end{equation}%

where $P^r$ is the raw point cloud and $\lambda$ balances the feature contribution of Point-NN and PointNet, which is set to 3 in our experiments.

\section{Dataset}

\subsection{Data Acquisition and Annotation}

The raw data were acquired from the actual test environment of polyester textile pilling, with each image containing approximately 200,000 points, covering an area of about $10\times20$mm in width and length. These images were captured using an SSZN8060 3D camera manufactured by Shenzhen DeepVision Intelligent Technology Co. Raw images were then annotated by an experienced industry professional using CloudCompare Software. The annotated images subsequently underwent a uniform downsampling process, reducing the number of sample points to 8192. While fewer downsampling points improve inference speed, researches \cite{d4srivastava2021exploiting,d5mohammadi2021pointview} indicate a potential trade-off with accuracy.

\begin{table*}[]
\setlength{\tabcolsep}{11pt}
\begin{center}
\begin{tabular*}{0.64\linewidth}{c|c|c|c}
\toprule
Method            & Input Points & OA(\%) & mAcc(\%) \\ 
\midrule
DGCNN \cite{c1wang2019dynamic}       & 8192         & 78.8   & 79.2   \\
DeepGCN \cite{c2li2019deepgcns}     & 8192         & 83.3   & 84.1   \\
PointNet \cite{c5qi2017pointnet}    & 8192         & 73.0   & 73.9   \\
PointNet++ \cite{c6qi2017et++}  & 8192         & 78.2   & 79.1   \\
PointNeXt-s \cite{c7qian} & 8192         & 85.0   & 86.1   \\
PointMLP \cite{d9ma2022rethinking}    & 8192         & 84.2   & 85.1   \\
Point-PN \cite{c8zhang2023}    & 8192         & 84.3   & 84.2   \\
PointGPT \cite{b10chen2024pointgpt}    & 8192         & 90.6   & 91.1   \\ \midrule
PointGPT+NN (Ours) & 8192         & \textbf{91.8}   & \textbf{92.2}   \\ 
\bottomrule
\end{tabular*}
\caption{Comparison of the effectiveness of the proposed method and other methods on the TextileNet8 dataset. The best results are highlighted in bold. OA is overall accuracy, and mAcc is mean accuracy per-class.} \label{tab:tab2}
\end{center}
\end{table*}

\begin{table*}[h]
\setlength{\tabcolsep}{11.0pt}
\centering
\begin{tabular*}{0.7\linewidth}{c|c|c|c|c|c}
\toprule
Mag. $\alpha$      & 100                                                   & 500                                                   & 1000                                                  & 2000                                                  & 3000                                                  \\ \midrule
Best OA(\%) & \begin{tabular}[c]{@{}c@{}}91.0\\ @$\beta$=300\end{tabular} & \begin{tabular}[c]{@{}c@{}}91.0\\ @$\beta$=100\end{tabular} & \begin{tabular}[c]{@{}c@{}}\textbf{91.8}\\ @$\beta$=100\end{tabular} & \begin{tabular}[c]{@{}c@{}}91.0\\ @$\beta$=400\end{tabular} & \begin{tabular}[c]{@{}c@{}}91.0\\ @$\beta$=200\end{tabular} \\ 
\bottomrule
\end{tabular*}
\caption{Results of search experiments for hyper-parameters $\alpha$ and $\beta$ of the trigonometric function in Point-NN. The second row of the table shows the best OA and the corresponding value of the hyper-parameter $\beta$. The best result is highlighted in bold.} \label{tab:tab3}
\end{table*}

\begin{table}[t]
\setlength{\tabcolsep}{9.0pt}
\centering
\begin{tabular}{c|c|c}
\toprule
Stage Num. & OA(\%) & mAcc(\%) \\ 
\midrule
1          & 90.6   & 90.5     \\
2          & \textbf{91.8}   & \textbf{92.2}     \\
3          & 89.5   & 89.5     \\
4          & 90.6   & 91.3     \\
5          & 89.9   & 90.1     \\ 
\bottomrule
\end{tabular}
\caption{Ablation study on the stage number of Multi-stage Hierarchy of Point-NN. The best results are highlighted in bold.} \label{tab:tab4}
\end{table}

\subsection{Statistics of TextileNet8}

The proposed textile classification dataset, TextileNet8, comprises 1335 3D point cloud images of polyester textiles. Following the relevant standard \cite{d7bar1988improved,d6goktepe2002fabric} for textile pilling classification, textile pilling can be grades range from 1 to 4.5 with intervals of 0.5. Accordingly, we categorize these images into eight grades (1-8), with the severity of textile pilling decreasing as the grade increases. Grade 1 has the fewest point cloud images at 150, while Grades 3 and 7 have the highest number of point cloud images at 180, and the rest of the Grades have 165 each. Table~1 presents the statistics of the TextileNet8 dataset and compares it with other related datasets. Notably, TextileNet8 stands out for being collected from an actual test environment, filling a gap in publicly available 3D point cloud datasets specifically related to industrial production. To the best of our knowledge, TextileNet8 represents the first publicly accessible dataset utilizing 3D point cloud data for eight-class textile pilling assessment. This dataset promises to advance research in textile pilling assessment based on 3D point cloud data.

\section{Experiments}

\subsection{Experimental Settings}
We conduct experiments using the TextileNet8 dataset as a benchmark to evaluate the performance of various models. The training and validation sets consisted of 1068 and 267 samples, respectively, with a ratio of 8:2, consistent with the popular ModelNet40 point cloud classification dataset \cite{e0wu20153d}. For a comprehensive analysis, we compare our proposed models with two graph-based methods, five MLP-based methods, and the vanilla PointGPT. All models are implemented using the PyTorch and trained on two Nvidia GTX A6000 GPUs. To accommodate GPU memory limitations, we select the PointGPT-S model with the fewest parameters as the base model, which is pre-trained on the ShapeNet dataset \cite{d8chang2015shapenet}. The batch size is 32 for training, a cosine learning rate (CosLR) scheduler is employed, and an AdamW optimizer uses an initial learning rate of 1e-4. The training epoch is 600 for optimal model convergence.

\subsection{Result Analysis}

Table 2 illustrates the performance of our proposed model and other models on the TextileNet8 dataset. As indicated in Table 2, our model achieves the best results in both OA and mAcc. Specifically, it outperforms the second-best model by 1.2\% in OA and 1.1\% in mAcc. Furthermore, the results indicates that graph-based point cloud processing methods exhibit mediocre performance, while MLP-based methods generally achieve an OA metric exceeding 84\%, which suggests that graph-based methods may not be suitable for textile pilling assessment tasks. Leveraging the powerful feature extraction capability of transformers, PointGPT and its improved version, PointGPT+NN, can achieve over 90\% OA and mACC, indicating their significant advantages over other methods for textile pilling grading.

\begin{table*}[t]
\setlength{\tabcolsep}{11.0pt}
\centering
\begin{tabular*}{0.83\linewidth}{c|c|c|c}
\toprule
Method                                & Data dimension & OA(\%) & mAcc(\%) \\ 
                                \midrule
Fusion on classification logits \cite{c8zhang2023} & 32x8                                 & 90.6   & 91.1     \\
Fusion on feature embedding     & 32x512x384                           & \textbf{91.8}   & \textbf{92.2}     \\ 
\bottomrule
\end{tabular*}
\caption{Comparison of the effectiveness of the proposed feature fusion strategy and the strategy proposed in the paper of Point-NN. The first number in the data dimension column is batch size. The best result is highlighted in bold.} \label{tab:tab5}
\end{table*}

\begin{table*}[t]
\setlength{\tabcolsep}{11.0pt}
\centering
\begin{tabular*}{0.64\linewidth}{c|c|c|c}
\toprule
Method      & Input Points & OA(\%) & mAcc(\%) \\ 
\midrule
DGCNN \cite{c1wang2019dynamic}       & 1024         & 92.9   & 90.2   \\
DeepGCN \cite{c2li2019deepgcns}     & 1024         & 93.6   & 90.9   \\
M-GCN \cite{c4hu2023m}       & 1024         & 93.1   & 90.1   \\
PointNet \cite{c5qi2017pointnet}    & 1024         & 89.2   & 86.0   \\
PointNet++ \cite{c6qi2017et++}  & 1024         & 92.7   & 90.1   \\
PointNeXt-s \cite{c7qian} & 1024         & 94.0   & 91.1   \\
PointMLP \cite{d9ma2022rethinking}    & 1024         & \textbf{94.5}   & \textbf{91.4}   \\
Point-PN \cite{c8zhang2023}    & 1024         & 93.8   & 91.2   \\
PointGPT \cite{b10chen2024pointgpt}    & 1024         & 94.0   & 91.1   \\ \midrule
PointGPT+NN (Ours) & 1024         & 94.2   & 91.3   \\ 
\bottomrule
\end{tabular*}
\caption{Comparison of the effectiveness of the proposed method and other methods on the ModelNet40 dataset. The best results are highlighted in bold.} \label{tab:tab6}
\end{table*}

\subsection{Ablation Analysis}

Table~3 presents the results of the ablation experiments conducted during the search for trigonometric function hyperparameters $\alpha$ and $\beta$. We test $\alpha$ values of 100, 500, 1000, 2000, and 3000, and $\beta$ values of 50, 100, 200, 300, and 400. Each $\alpha$ value is paired with each $\beta$ value, resulting in 25 combinations of $\alpha$ and $\beta$. The second row of Table~3 presents the highest OA achieved for each $\alpha$ value alongside its corresponding $\beta$ value. Notably, the model achieves its optimal OA when $\alpha$ is 1000 and $\beta$ is 100. Results from the table also indicate the best OA reaches 91.0\% across all combinations of each group, suggesting that the proposed PointGPT+NN model demonstrates robustness to these two hyperparameters.

Table 4 shows the results of tuning the number of stages of the multi-stage hierarchy of Point-NN. We vary the number of stages from 1 to 5. The smaller number of repetitions represents the more basic features obtained, and the original Point-NN model achieves its best results at a stage repetition number of 4 on the ModelNet40 dataset. As shown in Table 4, OA and mAcc reach their optimal values when the number of stages is 2, indicating that basic features are crucial for the model to get a good result from the TextileNet8 dataset.

Table~5 illustrates the results obtained from employing two different fusion strategies. The authors of Point-NN suggest the possibility of fusing the Point-NN's logit results with other models through linear interpolation to boost other models' performance. Therefore, we also attempt this approach. However, the results indicate no performance improvement. We meticulously examine the numerical outputs and conclude that its failure is likely due to the relatively small logit values produced by the Point-NN model and its significantly lower accuracy compared to the PointGPT model. It is important to note that using this fusion approach requires the usage of PMB of the Point-NN. Conversely, our strategy fusion in feature embedding ensures that PointGPT learns the overall features of the input raw point cloud, thereby enhancing model performance.

\subsection{Preformance on Other Dataset}

To validate the effectiveness of the proposed model, we conduct comparative experiments on the widely-used ModelNet40 dataset, which comprises 12,311 CAD point cloud images from 40 synthetic object categories. Table 6 shows the results. It is observed that although our proposed model does not achieve the best results, it achieves the second-best results in both OA and mAcc, which suggests that the proposed model has competitive performance on other larger datasets and is also valid to other point cloud classification tasks. The results also demonstrate the proposed model outperforms the vanilla PointGPT, indicating the feature fusion strategy is generalizable to improve the performance of the vanilla model.

\section{Conclusion}

In this paper, we present TextileNet8, the first publicly available eight-categories 3D point cloud dataset for textile pilling assessment. In addition, we effectively incorporate the point cloud features extracted from the adapted Point-NN model into the feature embedding of PointGPT, thus enhancing the performance of the vanilla PointGPT model on the TextileNet8 dataset. Comparative experimental results on TextileNet8 show that the proposed PointGPT+NN model achieves an OA of 91.8\% and a mAcc of 92.2\%. In addition, experimental results on ModelNet40 datasets show that the proposed PointGPT+NN model achieves better results than the vanilla model. In future work, we plan to collect more 3D point cloud data for textile filling evaluation and explore optimization methods to improve the performance of PointGPT+NN.

\bibliographystyle{kr}
\bibliography{template-paper}

\end{document}